\pdfoutput=1

\documentclass[11pt]{article}

\usepackage{acl}

\usepackage{times}
\usepackage{latexsym}
\usepackage{booktabs}
\usepackage{transparent}
\usepackage{multirow}
\usepackage{tabularx}
\usepackage{bold-extra}
\usepackage{ifthen}

\usepackage[T1]{fontenc}

\usepackage[utf8]{inputenc}

\usepackage{microtype}

\usepackage{inconsolata}

\usepackage{comment}
\usepackage{graphicx}

\definecolor{mydarkblue}{rgb}{0,0.08,0.45}
\definecolor{openaigreen}{RGB}{85, 180, 129}

\usepackage{csquotes}

\usepackage{hyperref}
\usepackage{CJKutf8}
\usepackage{awesomebox} 
\usepackage{bbding}
\usepackage[most]{tcolorbox}
\usepackage{fontawesome5}
\usepackage{tikz}
\usepackage[tikz]{bclogo}
\usepackage[framemethod=tikz]{mdframed}
\definecolor{bgblue}{RGB}{245,243,253}
\definecolor{ttblue}{RGB}{91,194,224}

\newtcolorbox{prompt}{
  colback=openaigreen!15, 
  colframe=gray!60, 
  boxrule=1pt, 
  arc=0pt, 
  boxsep=0pt, 
  left=6pt, 
  right=6pt, 
  top=6pt, 
  bottom=6pt, 
  enhanced, 
  fontupper=\small,
  grow to left by=-1mm,
  grow to right by=-1mm,
  overlay={
    \node[anchor=north east] at (frame.north east) {\transparent{0.2}\includegraphics[width=8mm]{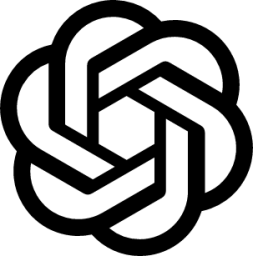}};
  }
}

\title{\textsc{JobSkape}: A Framework for Generating Synthetic Job Postings to Enhance Skill Matching}

\author{Antoine Magron\textsuperscript{*}\textsuperscript{1} \hspace{2em}
         Anna Dai\textsuperscript{*}\textsuperscript{1} \hspace{2em}
         Mike Zhang\textsuperscript{2} \\ 
         \textbf{Syrielle Montariol}\textsuperscript{1} \hspace{2em}
         \textbf{Antoine Bosselut}\textsuperscript{1}\\
  \textsuperscript{1}EPFL, Switzerland\\
  \textsuperscript{2}IT University of Copenhagen, Denmark\\
{\tt syrielle.montariol@epfl.ch} \hspace{2em} {\tt mikejj.zhang@gmail.com}}

\begin{document}
\maketitle

\begingroup\renewcommand\thefootnote{*}
\footnotetext{Equal contribution.}
\endgroup

\begin{abstract}
\looseness=-1
Recent approaches in skill matching, employing synthetic training data for classification or similarity model training, have shown promising results, reducing the need for time-consuming and expensive annotations. However, previous synthetic datasets have limitations, such as featuring only one skill per sentence and generally comprising short sentences. In this paper, we introduce \textsc{JobSkape}, a framework to generate synthetic data that tackles these limitations, specifically designed to enhance skill-to-taxonomy matching. Within this framework, we create \textsc{SkillSkape}, a comprehensive open-source synthetic dataset of job postings tailored for skill-matching tasks. We introduce several offline metrics that show that our dataset resembles real-world data.
Additionally, we present a multi-step pipeline for skill extraction and matching tasks using large language models (LLMs), benchmarking against known supervised methodologies. We outline that the downstream evaluation results on real-world data can beat baselines, underscoring its efficacy and adaptability. \footnote{Code and data available at \url{https://github.com/magantoine/JobSkape}}

\end{abstract}

\section{Introduction}
In the dynamic modern labor market, understanding job demands at scale is crucial for informed decision-making by policymakers, businesses, and other stakeholders. One way of measuring job market demand lies in \emph{skill matching}: the extraction and alignment of skills from job descriptions to their disambiguated forms (i.e., a knowledge base or taxonomy). This process facilitates the investigation of current labor market dynamics and the quantification of labor market demands, addressing the occupational skill matching problem. 

Regardless of their predictive effectiveness, supervised learning methods for skill matching require regularly collecting and annotating up-to-date data~\cite{zhang-etal-2022-kompetencer}, a process that is both expensive and time-consuming. Synthetic data circumvents the need for such costly annotations.
However, despite efforts in generating synthetic training data~\cite{DBLP:conf/hr-recsys/ClavieS23, DBLP:journals/corr/abs-2307-10778} and real-world benchmarks~\cite{zhang-etal-2022-skillspan, decorte2022design}, challenges like incoherent sentences and over-simplified setups exist in existing datasets. To address these shortcomings, we introduce \textsc{JobSkape}, a framework for generating realistic skill matching datasets that can be used for training and benchmarking. 

\textsc{JobSkape} facilitates the creation of diverse labeled textual datasets that align closely with actual job postings, ensuring cleaner and more coherent data. We demonstrate its practical application by generating \textsc{SkillSkape}, a large-scale dataset linking coherent sets of skills to corresponding job descriptions. \textsc{JobSkape} uses generative large language models (LLMs) to curate meaningful skill combinations and generate appropriate job descriptions containing these combinations. A self-refinement step using LLMs \cite{DBLP:journals/corr/abs-2303-17651} ensures label quality in the refined \textsc{SkillSkape} dataset, assessed through offline metrics.
Finally, we challenge traditional supervised skill matching methods with an LLM-based, in-context learning (ICL) pipeline, to circumvent re-training the supervised model given new data. We evaluate skill matching performance on our synthetic dataset and real-world annotated data~\citep{decorte2022design}, comparing our proposed extraction and matching pipeline with supervised matching models trained on our dataset as well as previous generation attempts ~\citep{DBLP:journals/corr/abs-2307-10778} in a controlled setting.

\paragraph{Contributions.} In this work, we contribute the following: (1) we propose \textsc{JobSkape}, a framework for generating a synthetic dataset of job descriptions for skill matching with existing skill taxonomies, (2) using our framework, we release a synthetic train and evaluation dataset (\textsc{SkillSkape}) for skill matching, (3) we show that \textsc{SkillSkape} has higher textual quality measured in perplexity and implicitness compared to previous synthetic datasets, (4) lastly, we introduce an ICL-based approach to extract and match skills from job descriptions to a taxonomy and show that this method can outperform supervised baselines on real-world benchmarks.

\section{Related Work}

\paragraph{Synthetic Data Generation.} Traditional synthetic data generation relies on language models, where a generator model is trained on an existing dataset and then employed to generate new data \citep{mohapatra-mohapatra-2022-sentiment, kumar2020data}. More recent unsupervised methods, such as \citet{wang2021towards}, leverage pre-trained language models like GPT-3~\cite{brown2020language} without the need for explicit supervision. Other examples include \citet{ye-etal-2022-zerogen, gao2023self}, who use carefully designed prompts for data generation. \citet{honovich2022unnatural} generate synthetic instructions for fine-tuning large language models, while \citet{shao2023synthetic} create synthetic demonstrations to enhance the performance of prompting LLMs. 

\paragraph{Synthetic Data for Job Postings.} In the job market domain, \citet{DBLP:journals/corr/abs-2307-10778, DBLP:conf/hr-recsys/ClavieS23} both employ GPT-3.5/4 to generate synthetic training data for skill matching. Specifically, \citet{DBLP:journals/corr/abs-2307-10778} prompt GPT-4 to generate ten examples for each ESCO skill, while \citet{DBLP:conf/hr-recsys/ClavieS23} use GPT-3.5 to generate 40 examples for each ESCO skill. In this work, we compare our dataset with the one from \citet{DBLP:journals/corr/abs-2307-10778}, referred to hereafter as the \textsc{Decorte} dataset.

\paragraph{Skill Matching.} Earlier works focus on standardizing skills through matching with taxonomies. For supervised methods, \citet{gnehm-etal-2022-fine} extract skills from Swiss-German job descriptions and match them with the ESCO taxonomy in a two-step process. \citet{zhang-etal-2022-kompetencer} assume pre-extracted skills and classify spans into their respective taxonomy codes using multiclass classification. \citet{decorte2022design} use distant supervision with the ESCO taxonomy to obtain labels, employ binary classifiers for each ESCO skill and enhance training through negative sampling strategies. 
\citet{DBLP:journals/corr/abs-2307-10778, DBLP:conf/hr-recsys/ClavieS23} employ LLMs for skill matching with ESCO. \citet{DBLP:journals/corr/abs-2307-10778} generate a synthetic training set using GPT-3.5 and optimize a bi-encoder through contrastive training for matching. \citet{DBLP:conf/hr-recsys/ClavieS23} use a similar approach, generating synthetic training data and employing a linear classifier for each skill with a negative sampling strategy. Additionally, they use sentence embedders \cite{reimers-gurevych-2019-sentence} to measure the similarity between extracted skills and ESCO.

\section{The \textsc{JobSkape} Framework}
Our goal is to create a synthetic dataset comprising job posting sentences associated with lists of skills from a taxonomy that closely aligns with real-world job posting sentences. We initiate the process by generating combinations of skills, derived from a given taxonomy, that are likely to coexist in a job description. Leveraging LLMs and refinement techniques, we produce diverse, realistic, and accurate job description sentences. To evaluate the quality of our synthetic data generation, we define a set of offline metrics and compare the generated sentences with real job postings.

\subsection{The Label Space}\label{sec:label-space}
In this study, we use the European Skills, Competences, Qualifications, and Occupations (ESCO; \citealp{le2014esco}) taxonomy as the label space. ESCO comprises 13,890 competencies categorized into \emph{Skill}, \emph{Knowledge}, and \emph{Attitudes}. Knowledge, according to ESCO, involves assimilating information through learning, encompassing facts, principles, theories, and practices in a specific field of work or study.\footnote{\url{https://ec.europa.eu/esco/portal/escopedia/Knowledge}} For example, acquiring proficiency in the Python programming language through learning represents a \emph{knowledge} component, classified as a \emph{hard skill}. Conversely, the application of this knowledge to perform tasks is considered a \emph{skill} component, defined by ESCO as the ability to apply knowledge and use know-how to accomplish tasks and solve problems.\footnote{\url{https://ec.europa.eu/esco/portal/escopedia/Skill}} For the synthetic sentence generation task at hand, we do not distinguish between skill and knowledge components.

Our synthetic dataset creation framework generates sentences containing multiple skills listed in the ESCO taxonomy. To reduce the data generation cost (and to facilitate a fair comparison with prior work), we use the same subset of 514 ESCO skills used in \textsc{SkillSpan-M}, an annotated set of real-world job postings.

\subsection{Formal Approach}

Previous efforts (\citealp{DBLP:journals/corr/abs-2307-10778, DBLP:conf/hr-recsys/ClavieS23}) focused on generating synthetic training sentences with a single skill. In contrast, we advocate for sentences containing multiple relevant skills to resemble sentences from real job postings. We initiate the process by creating combinations of skills, guided by three main conditions: 
\begin{enumerate}
    \item \textbf{Varying Lengths of Skill Combination}: Recognizing the heterogeneity in real-world job postings, we incorporate varying numbers of skills per sentence. By doing so, our sentences will show higher diversity similar to real job advertisements.
    \item \textbf{Semantic Closeness in Skill Pairing}: In real job postings, skills that are mentioned in the same sentence are often related to each other. Aligning with the logical grouping of skills, we construct more realistic and contextually coherent sentences.
    \item \textbf{Minimum Skill Representation}: While our dataset aims to reflect the real-world frequency of skill occurrences, we also want to ensure that each skill appears enough times for training. This guarantees that even less common skills are adequately represented, creating a more effective training dataset.
\end{enumerate}

To achieve variety, we introduce two distributions, $\mathcal{N}$ (distribution of combination size) and $\mathcal{F}$ (distribution describing skill frequency in job postings, akin to skill popularity). We iteratively process skills $s_i \in \mathcal{S}$, the set of skills in our taxonomy, ensuring each skill has the same minimum number of samples. For each skill, we identify its $k$ nearest neighbors $\{s_j'\}_{j=1}^k$ based on cosine similarity between embeddings obtained from JobBERT, a language model fine-tuned on domain-specific data \cite{zhang-etal-2022-skillspan}. Neighbors with a similarity above threshold $T$ are retained, forming the set of nearest neighbors $\mathcal{S}_i$:
\begin{equation}    
\mathcal{S}_i = \{ s_j : s_j \in \{s_i'\}_{i=1}^k \wedge \text{sim}(s_j, s_i) \} T\}.
\end{equation}

This set is used for skill combination selection. We draw a sample size $n$ from distribution $\mathcal{N}$, setting the combination size to $min(n, |\mathcal{S}_i|)$. However, sampling skills directly from $\mathcal{S}_i$ is not straightforward. For instance, in analyzing the top nearest neighbors of \textit{SQL}, a frequently occurring skill in job postings, we find \textit{THC Hydra}, which is much less common. To accurately replicate the real-world frequency distribution of skills in our synthetic dataset, we adjust sampling probabilities to reflect actual skill popularity. 
Hence, we introduce distribution $\mathcal{F}$ to compute the probability of selection over $\mathcal{S}_i$ using softmax:
\begin{equation}
\mathbf{P}(s_j) = \frac{e^{\mathbf{P}_{\mathcal{F}}(s_j) }}{\sum_{l=1}^k e^{\mathbf{P}_{\mathcal{F}}(s_l)}}.
\end{equation}

\noindent where $\mathbf{P}_{\mathcal{F}}(s_i)$ is the popularity of the skill $s_i$. We then select $min(n, |\mathcal{S}_i|)$ skills from $\mathcal{S}_i$ using the computed probability distribution.

For dataset creation, we form skill combinations from our ESCO subset, with $\mathcal{N}$ set to $U(1, 5)$. We employ JobBERT to obtain domain-specific embeddings for job descriptions and skills. The distribution $\mathcal{F}$ is computed as the average of standardized negative perplexities across sentences generated with GPT-2~\citep{radford2019language}. These sentences include variations like \texttt{``I want a job that involves \{skill\}''}, \texttt{``For my job, I want to learn [to] \{skills\}''}, \texttt{``At my job, my main is skill is [to] \{skill\}''}, ensuring grammatical correctness. We set a similarity threshold $T$ to $0.83$ to be closer to SkillSpan distribution, using $k = 20$.

\subsection{Prompt Tuning for Generation}

Given a skill combination, we generate synthetic job description sentences. A candidate for this hypothetical job would need to be proficient to some extent in each of these skills. We use GPT-3.5 as the text generator. We describe two types of generations:

\begin{itemize}
    \itemsep0em
    \item \textbf{Dense}: For a combination of four or less skills we generate a short job description of at most one sentence. This is done to minimize the number of hallucinated skills that could appear when generating a long job description with a small set of skills.
    \item \textbf{Sparse}: For a combination of more than four skills, we generate a job description paragraph containing multiple sentences. The information is more ``sparse''.
\end{itemize}

Our prompt follows \citet{DBLP:conf/hr-recsys/ClavieS23}, it is used to make the mentions of the skills as implicit as possible (i.e., skill does not have an exact string match in the text). We further enhance the diversity by prompting the model to vary the openings of the descriptions and avoid the examples starting with \texttt{``We are looking''} or \texttt{``We are searching''} (see Appendix \ref{sec:datagen-positive-samples}). 

We add additional instructions to the prompt to reduce ambiguity. To each prompt, we add a list of synonyms of each inputted skill that are in the taxonomy and instruct the model to not refer to \textit{SQL} as \textit{MySQL} since \textit{MySQL} is a separate skill in the taxonomy. Each skill is also given along with their respective definitions to give more context to the model and avoid miscomprehension.

\begin{figure*}[t]
    \centering
    \includegraphics[width=.9\linewidth]{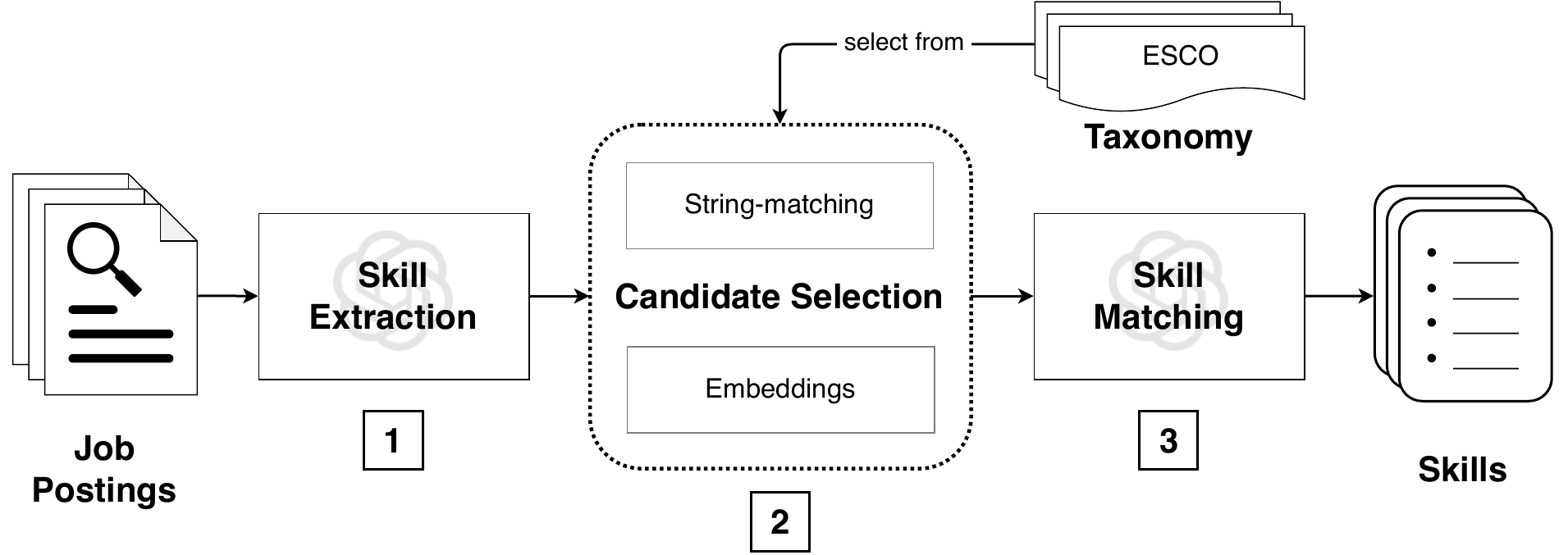}
    \caption{\textbf{Three-step Skill Extraction and Matching Pipeline}. We show our in-context learning pipeline for end-to-end skill matching. We use an LLM to extract skills from job ads, then do candidate selection using heuristics, and last, do skill matching with a constrained taxonomy.}
    \label{fig:pipeline}
\end{figure*}

\subsection{Refinement of \textsc{SkillSkape}}\label{sec:refinement}

At this step, the dataset comprises exclusively of positive samples, which means that every generated sentence contains at least one associated skill. To train a supervised classifier for real job descriptions, negative samples -- sentences containing unknown skills or no skills -- are required. 
To create negative samples with unknown skills, we apply the same generation method as positive samples but draw from a broader pool of skills. This pool includes skills that are not in our selected list but known in the wider skill universe. We also generate negative samples containing no skills to represent sentences in real-life job ads that do not mention skills required from the candidate. To do so, we use two separate prompts to generate (1) sentences describing the company: its reach, domain, location, et cetera and (2) sentences detailing the salary and perks of a job (see Appendix \ref{sec:datagen-negative-samples}). To guide the model in the generation, we provide two demonstrations.
For the \textsc{SkillSkape} dataset, we generate 500 negative samples with unknown skills as well as 500 additional negative samples containing no skills. 

We then apply self-refinement \cite{DBLP:journals/corr/abs-2303-17651}, involving feeding the generated sentences back into the same model for feedback. The model is asked to extract skills using the pipeline described in Appendix \ref{sec:datagen-initial-prompt}, matching them with the taxonomy. We compare the generated set of skills with the gold set of skills, adding to the gold list all skills that were found in the sentence. We do this because the LLM can extrapolate during sentence generation, thereby adding related skills on top of the original list that was fed to it. The list of skills, along with their associated spans in the sentences, is filtered to include only pairs of skills and spans that have a cosine similarity above a specified threshold. For this refined dataset version, we use JobBERT as a span encoder, and the threshold is empirically set to $0.7$ cosine similarity. One of the main reasons for this low similarity is the LLM not reflecting accurately the gold skill during the sentence generation step. In that case, we wish to remove that skill from the gold set of skills associated with the sentence.

\paragraph{Span Extraction.} We use GPT-3.5 to label skill sequences in the sentences. Each mention, whether implicit or explicit, is surrounded by \texttt{@@} and \texttt{\#\#} following~\citet{wang2023gpt}. In case the language model fails to label the span, it is asked to self-correct, as outlined in Appendix \ref{sec:datagen-refining-shots}. 
We showcase two examples extracted from the training set of the refined \textsc{SkillSkape} dataset.

\paragraph{Positive example.} This 28-word example, average in length for our dataset, contains three key skills as annotated spans required of the applicant.
\begin{quote}
    \vspace{-2em}
    \itemsep0em
    \item[] \texttt{\textbf{Sentence}: The ideal candidate will effectively @@engage with upper-level management\#\#, @@maintain strong communication channels with key stakeholders\#\#, and @@collaborate with peers\#\# to ensure seamless coordination throughout the organization.}
    \item[] \texttt{\textbf{Label}: 'liaise with managers', 'communicate with stakeholders', 'liaise with colleagues'}
\end{quote}

\noindent In the example, the inter-skill similarity is high, showcasing the efficiency of the skill combination selection method.

\paragraph{Negative example.} This sentence mentions information about the hiring company instead of the job itself, and therefore, contains no skills.
\begin{quote}
    \vspace{-2em}
    \itemsep0em
    \item[] \texttt{\textbf{Sentence}: Embrace a challenging and fulfilling career with us, where your hard work is recognized through a salary range of \$80,000 to \$90,000, reflecting our appreciation for your contributions.}
    \item[] \texttt{\textbf{Label}: NO LABEL}
\end{quote}

\begin{table}[t]
    \centering
    \resizebox{\linewidth}{!}{
    \begin{tabular}{llcccc}
        \toprule
           &  & Avg. & & Avg. & \\ 

          Dataset &  Split  & \# Skills & \% \texttt{UNK}s & \# Words & \# Samples\\ 
        \midrule
        \multirow{2}{*}{\textsc{SkillSpan-M}}   & Dev.   & 2.0   & 47.0  &   15.0   & 178   \\  
                  & Test     & 1.9    & 47.3  & 16.3   & 751  \\  
        \midrule
        \textsc{Decorte}  & Train  & 1.0     & 0.2   & 15.7    & 5,120 \\ 

        \midrule
                  & Train    & 2.6   & 7.9    &  28.2          & 6,352 \\
        \textsc{SkillSkape}  & Dev.  & 2.1  & 8.3   &  27.8   & 1,316 \\ 
                  & Test     & 2.6   & 8.4    &28.1         & 1,272 \\          
        \bottomrule
    \end{tabular}}
    \caption{\textbf{Datasets' Statistics.} Average \# skills and words refer to the average per sample (job posting sentence(s) and \% \texttt{UNK}s refer to the percentage of skill labels are under the unknown \texttt{UNK} label.}
    \label{tab:ds_stats}
\end{table}

\subsection{Summary and Comparison}

The final version of \textsc{SkillSkape} has 8940 samples, split into training, development and test sets ($\sim$ 70-15-15 split). We provide several descriptive statistics in Table~\ref{tab:ds_stats}. To assess the quality of our generations, we compare the generated dataset \textsc{SkillSkape} with two other datasets from the literature: (1) a manually annotated benchmark, created by~\citet{decorte2022design}, based on the \textsc{SkillSpan-M(atch)} dataset~\cite{zhang-etal-2022-skillspan}, which contains over 14.5K job posting sentences scraped from various sources, 
and (2) the \textsc{Decorte} dataset \cite{DBLP:journals/corr/abs-2307-10778}, synthetically generated from ESCO using GPT-4.
By design, we created \textsc{SkillSkape} to cover the same label space as \textsc{SkillSpan-M}, which has only a development and a test set. In that dataset, two labels are used to indicate skills without an adapted label in the taxonomy: \texttt{UNDERSPECIFIED} and \texttt{LABEL NOT PRESENT}. We map these to the \texttt{UNK} label used in \textsc{SkillSkape}.
\textsc{Decorte} associates ten synthetic sentences to each skill in the ESCO taxonomy. It is only used as training data. It covers all of ESCO (13.9K skills), but we restrict it to sentences with skills occurring in \textsc{SkillSpan-M}, leading to 5,120 samples (we add 10 random \texttt{UNK} sentences).

\section{Experimental Setup}

In this section, we introduce several benchmarks for skill matching tasks. We train a supervised multi-label classifier and present an LLM-based approach with in-context learning.

\subsection{Supervised Multi-label Classifier}
For the supervised baseline, we use a pre-trained BERT$_{base\_uncased}$ model~\cite{devlin-etal-2019-bert} to extract contextualized embeddings from the input text $t = \{w_1, w_2, ..., w_n\}$. These embeddings are then input into a multi-label classifier with a sigmoid activation applied independently to each output logit. Let $y = \{y_1, y_2, ..., y_k\}$ represent binary labels for the $k$ classes. The model predicts the labels using:

\begin{equation}
\hat{y}_i = \sigma(f_i(\text{BERT}(\boldsymbol{t}))), 
\end{equation}

\noindent where $f_i(\cdot)$ is a function that maps the output embeddings from BERT to a logit for class $i$, $\sigma(\cdot)$ is the sigmoid activation, and $\hat{y}_i$ is the predicted probability for class $i$. The probability threshold can be tuned; we empirically found that 0.2 works well for this task.

We train the BERT$_{base}$ model for 100 epochs with a learning rate of $3 \times 10^{-5}$ and select the best-performing epoch. We use a batch size of 16 and a maximum sequence length of 128. The model is trained for five different seeds. 

\subsection{In-context Learning with LLMs}

We leverage an LLM to match skills in synthetic job posting sentences to the ESCO taxonomy. This pipeline has three steps, visualized in Figure~\ref{fig:pipeline}: 1) \textbf{skill extraction} from the sentence, 2) \textbf{candidate selection} from the taxonomy, and 3) \textbf{skill matching} to the list of candidates. Here, we first extract relevant skills using LLM-prompting, pre-select viable candidates from our taxonomy, and then match the skills to candidates in the taxonomy through LLM-prompting again. We adopt a three-step approach to overcome the limited context window of LLMs, specifically 4K for \texttt{GPT-3.5-turbo}~\cite{openaigptdocs}, which makes feeding large taxonomies directly to the model impractical.

\paragraph{(1) Skill Extraction.} 
For each job posting, the LLM identifies key skills and tasks within the job ad while omitting irrelevant information. The LLM is directed to respond by repeating the sentence and tagging the skills by surrounding them with \texttt{@@} and \texttt{\#\#}, following \citet{wang2023gpt}, as shown in our prompt below.

\begin{quote}
    \vspace{-2em}
    \itemsep0em
    \item[] \texttt{\textbf{System}: You are an expert human resource manager. You need to analyse skills in a job posting.}
    \item[] \texttt{\textbf{Instruction}: You are an expert human resource manager. You are given an extract from a job description. Highlight all the skills, competencies and tasks that are required from the candidate applying for the job, by surrounding them with tags @@ and \#\#. Make sure you don't highlight job titles, nor elements related to the company and not to the job itself. Make sure to rewrite the sentence with all the tags.}
    \item[] \texttt{\{Demontrations\}}
    \item[] \texttt{Sentence: \{Sentence\}}
    \item[] \texttt{Answer: }
\end{quote}

We provide seven demonstrations in a few-shot setting to assist the model in understanding the task and following the instructions. To select few-shot examples, we use $k$NN retrieval from a training set composed of sentences along with their spans and labels~\cite{liu-etal-2022-makes}. The closest samples from our dataset are selected as few-shot examples (see Appendix \ref{sec:app-prompt}).
Finally, We process the output by extracting the tagged sections as skills. 

\paragraph{(2) Candidate Selection.}
Matching extracted skills with skills defined in the taxonomy is crucial. Each skill in the taxonomy is associated with a tiered structure of names, each providing different levels of detail, and a definition. To provide richer context to the model, we concatenate the most detailed (or granular) name of the skill with its definition. We use two methods for pre-selecting viable candidates from the taxonomy for each skill:

\begin{itemize}
    \itemsep0em
    \item \texttt{\textbf{rule-based}}: Through string matching, we seek full or approximate matches of the extracted skill within the taxonomy. If the exact string of the extracted skill is present in the name or definition of a skill in the taxonomy, it is considered a good candidate for a match. We randomly select five entries if more than five candidates are found. If the exact strings do not match, we calculate the \texttt{token\_set\_ratio} using TheFuzz,\footnote{\url{https://github.com/seatgeek/thefuzz}} a similarity score based on Levenshtein's distance~\cite{levenshtein1966binary}. The top five candidates with the highest scores are chosen.
    \item \texttt{\textbf{embedding-based}}: Using a pre-trained language model (JobBERT;~\citealp{zhang-etal-2022-skillspan}), we compare the extracted skills with taxonomy entries. We obtain the contextualized embeddings of the extracted skill by embedding the sentences and averaging the vector representation of the tokens of the extracted skill. These embeddings are then compared to the representation of each skill in the ESCO taxonomy if the extracted skill is a substring of the sentences it was extracted from. Otherwise, the embedding of the extracted skill itself is compared to the skills in the taxonomy. The top five most similar candidates are selected based on cosine similarity.
\end{itemize}

\noindent
While effective, the rule-based method may miss synonyms and context. On the other hand, the embedding-based method addresses the limitations of the rule-based method but risks selecting contextually similar yet factually dissimilar candidates (e.g., software vs. hardware). Therefore, we adopt a hybrid approach, retaining candidates from both rule-based and embedding-based methods.

\paragraph{(3) Skill Matching.}

The final step involves matching extracted skills to one of the selected candidate skills. We present the LLM with formatted candidates as options, 
and request the best match, resembling a ranking task. The model outputs the most fitting option as a matched skill or provides no match if none are found.  To assist the model without overloading the prompt, we provide a one-shot example in the following format:
\begin{itemize}
    \itemsep0em
    \item[] \texttt{Sentence: \{generated sentence\}}
    \item[] \texttt{Skill: \{extracted span\}}
    \item[] \texttt{A: \{candidate 1\}} 
    \item[] \texttt{...} 
    \item[] \texttt{J: \{candidate 10\}} 
    \item[] \texttt{Answer: \{selected candidate\}} 
\end{itemize}

The full prompt with one-shot demontration can be found in Appendix \ref{sec:matching-prompt}.

\noindent To conduct the experiments for in-context learning with LLM, we retrieve demonstrations from the training set to provide examples for both the extraction and matching steps. We conduct an ablation study on SkillSpan's validation set to select the best number of shots for both tasks. Experiments are described in Appendix~\ref{sec:app-ablation}, in Table~\ref{tab:app-ablation_ndem}, and Figure~\ref{fig:app-ablation_ICL}. The matching step is performed with 10 candidates using the mixed setting (5 embedding candidates and 5 string matching candidates). The best setting uses 7 demonstrations for the extraction step and one demonstration for the matching step. Matching step demonstrations have a large number of tokens due to the list of candidates along with their definitions, which can explain the decreased performance associated with adding more demonstrations.

\begin{table}[t]
    \centering
    \resizebox{\linewidth}{!}{
    \begin{tabular}{lrrr}
          \toprule
                                & Perplexity ($\downarrow$)      & S2SIM ($\uparrow$)                 & Explicitness (\%, $\downarrow$) \\ \midrule
        \textsc{SkillSpan-M}    & 178.2         & 0.662           &  5.0  \\ 
        \textsc{Decorte}        & 65.1          & 0.739  &      22.4\\ 
        \textsc{SkillSkape}      & 44.3 & 0.744          &  6.9   \\ \bottomrule
    \end{tabular}
    }
    \caption{\textbf{Offline Metrics.} We show the offline metrics as described in Subsection~\ref{subsec:offline}. ($\uparrow$) indicates higher the better, ($\downarrow$) indicates lower is better.} 
    \label{tab:off_metrics}
\end{table}

\begin{table*}[t]
\centering
\resizebox{.9\linewidth}{!}{
\begin{tabular}{lrr c rr}
\toprule
&\multicolumn{2}{c}{\textbf{Supervised}} & & \multicolumn{2}{c}{\textbf{Few-Shot ICL}} \\
\cline{2-3} \cline{5-6}
$\downarrow$Train / Test$\rightarrow$ & \textsc{SkillSkape}  &\textsc{SkillSpan-M}  & & \textsc{SkillSkape}  & \textsc{SkillSpan-M}   \\
\midrule
\textsc{Decorte}           & 28.0 $\pm$ 0.8  & 23.0 $\pm$ 0.7   & & 36.8 $\pm$ 0.2 & 26.9 $\pm$ 0.5  \\
\textsc{SkillSkape}       & \textbf{68.0 $\pm$ 0.5} & 22.2 $\pm$ 0.9   & & \textbf{37.6 $\pm$ 0.2} & 26.9 $\pm$ 0.3 \\
Both            & 67.2 $\pm$ 1.0    & \textbf{26.1 $\pm$ 1.2}  & & \textbf{37.6 $\pm$ 0.2} & \textbf{27.3 $\pm$ 0.4} \\         
\bottomrule
\end{tabular}
}
\caption{\textbf{Supervised and Few-Shot ICL Results}. \emph{Both} indicates the concatenation of \textsc{Decorte} and \textsc{SkillSkape}. The scores are micro-$F_1$.}
\label{tab:test_un_supervised}
\end{table*}

\subsection{Offline Quality Metrics}\label{subsec:offline}

We design a set of metrics to evaluate the quality and diversity of the data at hand. Our intention is not to mirror metrics of \textsc{SkillSpan-M}, which is untidy by nature of scraped data, but to produce high-quality training data for downstream skill matching tasks. 
\begin{enumerate}
    \itemsep0em
    \item First, we consider \textbf{Perplexity}, i.e., how realistic the data is from the point of view of a language model. We compute the perplexity of each of the sentences using GPT-2~\cite{radford2019language}, where lower is better.
    \item Second, we consider \textbf{Skill-Sentence Similarity (S2SIM)}, the average cosine similarity between a skill and the associated sentence. The higher this metric, the closer the generated sentence will be semantically close to the associated skills. We aim to maximize this metric. The embeddings are computed using JobBERT, and BERT model fine-tuned on English job postings with the masked language modeling objective.
    \item Finally, we measure \textbf{Explicitness} by counting the number of entities that appear exactly in the sample, using string matching. 
\end{enumerate}

Table \ref{tab:off_metrics} shows offline metrics for \textsc{SkillSpan-M}, \textsc{Decorte}, and \textsc{SkillSkape}. \textsc{SkillSkape} has a lower perplexity, and outperforms \textsc{SkillSpan} and \textsc{Decorte} in terms of S2SIM. The main reason for \textsc{SkillSpan-M}'s low skill-sentence similarity is its noisiness, leading to sentences often being cut mid-way and lacking coherence.
Around 7\% of \textsc{SkillSkape} skills are fully explicit (the label can be found exactly in the sentence), much closer to \textsc{SkillSpan-M} than \textsc{Decorte}. A higher explicitness leads to an easier task; a skill matching model needs to be trained on enough implicit examples to allow it to generalize to implicit skills.

Overall, \textsc{SkillSkape} demonstrates similar statistics in perplexity and S2SIM characteristics as \textsc{Decorte}. However, it notably exhibits a significant ($3\times$) enhancement in implicitly representing skills within each sentence.

\section{Results and Analysis}

To assess the label refinement method (Section \ref{sec:refinement}), we apply it to the development set of the \textsc{SkillSpan-M} benchmark that has annotated skills and associated spans. $40\%$ of our extracted spans match exactly with the annotated span. $60\%$ of our extracted spans are either a perfect match or contain the annotated span. In general, extracted spans have a Jaccard similarity of $62\%$ with the annotated spans.

\subsection{Supervised vs. Few-shot ICL Matching}

In Table~\ref{tab:test_un_supervised}, we show the results of the skill matching task on the \textsc{SkillSpan-M} test set and \textsc{SkillSkape} test set. We compare the performance of supervised and in-context learning methods \textit{trained} on the \textsc{Decorte} training set, \textsc{SkillSkape} training set, or the concatenation of both.
The supervised approach uses training data to train a supervised multi-label classifier, whereas the few-shot ICL approach uses it as a demonstration pool to retrieve $k$NN demonstrations. For simplicity, we refer to both training and few-shot learning with demonstrations as \textit{training} in the remainder of this section.

Comparing across supervised and few-shot in-context learning settings, we observe that both supervised and ICL approaches achieve higher performance on both real-world data (\textsc{SkillSpan-M}) and our synthetic dataset (\textsc{SkillSkape}) when trained on \textsc{SkillSkape} training set or a combination of \textsc{SkillSkape} and \textsc{Decorte} training sets. This increase in matching performance is likely due to a higher textual diversity in the \textsc{SkillSkape} dataset. Across both matching approaches, we also observe that training on both \textsc{SkillSkape} and \textsc{Decorte} consistently achieves the highest test $F_1$ scores on real-world data. However, the difference in performance is greater for the supervised approach than ICL, highlighting that the ability to generate high-quality data is most impactful for supervised approaches.

Additionally, we observe an interesting result in the large difference between the supervised and ICL performance on the \textsc{SkillSkape} test set, 68.0/67.2 and 37.6 micro-$F_1$ respectively when trained on \textsc{SkillSkape} or a combination of \textsc{SkillSkape} and \textsc{Decorte}. We suspect that this difference could largely be due the characteristics of our training and test data. Supervised models tend to perform well when the training and test data follow the same distributions.
In contrast, the few-shot ICL method consistently outperforms than the supervised approach on the \textsc{SkillSpan-M} test set. Given the minimal tuning required for the ICL method, the ICL approach can be better suited to flexibly handle messy real-world data.
These results suggest that, for use cases when we have a sample of annotated data from the same distribution as the data we want to predict, we can combine it with synthetic training data and leverage supervised models. Otherwise, the in-context learning approach is less dependent on the training data.

In Appendix Table~\ref{tab:quali}, we show several qualitative examples of predictions of both the multi-label classifier and LLM. Several noticeable patterns are underprediction for the multi-label classifier and overprediction of the LLM. Additionally, we notice that the predictions of both models are rather close ``semantically'' to the gold labels, but are deemed incorrect by the evaluation.

In summary, the results underscore the significance of both the quantity and diversity of training data in the development of effective skill matching dataset generators.

\subsection{Effect of In-context Demonstrations}

We evaluate the sensitivity of our method to the number and candidate selection methodology of in-context learning examples. 

\paragraph{Demonstrations.}
We perform an ablation study on the number of demonstrations for both skill extraction and matching. Results in Appendix Table \ref{tab:app-ablation_ndem} show that 7 shots for extraction with 1 shot for matching leads to the best performance.

\paragraph{Candidate Selection.} 
Candidate selection using the hybrid method for $n=5$ candidates from each of the rules- and embedding-based methods (i.e., 10 candidates in Figure \ref{fig:num_cand})  presents the best trade-off between performance and computational cost. While we do observe a higher $F_1$ score as we increase the number of candidates, the increase in performance appears to be marginal while it would more than double the number of input tokens.

Finally, an ablation study on the matching step of the pipeline (see Appendix\ref{sec:app-ablation-matching}) shows that directly selecting the top-1 candidate (rule-based) as skill prediction lags behind the performance of using GPT-3.5 as a re-ranker by around 8\% $F_1$.

\subsection{Effect of Sentence Length}

The sentence length distribution is heavily skewed toward shorter sentences in the \textsc{Decorte} and \textsc{SkillSpan-M} test sets, with 50\% of sentences being 13--19 words in \textsc{Decorte} and 7--20 words in \textsc{SkillSpan-M}. In contrast, 50\% of the sentences in \textsc{SkillSkape} are between 23--33 words (see Figure \ref{fig:app-sent-len} in Appendix for a visualization of the length distribution for each dataset).

When splitting the \textsc{SkillSpan-M} test set into two equal-sized sets depending on the size of the sequence (less than 12 words, or more than 12 words), training on \textsc{Decorte} leads to slightly higher performance than \textsc{SkillSkape} for shorter sentences (0.26 vs. 0.24 $F_1$). For longer sentences, however, \textsc{SkillSkape} reaches an $F_1$-score of 0.18 while Decorte's $F_1$ is 0.17.

\section{Conclusion}

We introduce \textsc{JobSkape}, a general framework for generating synthetic job posting sentences for skill matching. Using our framework, we release \textsc{SkillSkape} a large dataset of synthetic job posting sentences labeled with ESCO skills. Our analysis shows that \textsc{SkillSkape} contains more implicit skills, has longer sentences, and is overall closer to real-world data, compared to alternative synthetic dataset from the literature.
Using our dataset, we conducted several skill matching experiments by training a supervised multi-label classifier and using in-context learning with an LLM, and showed that both methods achieved comparable results when evaluated on real-world data ($F_1$ of 26.1 and 27.3 respectively). 
Furthermore, we note that the potential applications of \textsc{JobSkape} extend beyond its current scope. Its application in creating synthetic CVs, for instance, can enhance job matching algorithms and facilitate skill-gap analysis in various industries. The framework's adaptability to different skill taxonomies also opens up possibilities for use across multiple sectors. While promising, these extended applications require further exploration to fully assess their impact.

\section{Limitations}

\paragraph{Closed model.} One of the primary limitations comes from our use of Large Language Models (LLMs) that are closed. This restricts our ability to understand, modify, or customize the underlying mechanisms of these models. The closed nature of the LLMs used in our study also limits the transparency, adaptability, and reproducibility of our system.

\paragraph{English only.} Our method is limited to processing and understanding English language content. This language-specific focus narrows the scope of our system's applicability, excluding non-English speaking demographics.

\paragraph{Bias inherited from LLMs.} Another significant limitation is the potential bias inherited from the LLMs. Since these models are trained on large datasets that may contain biases, there is a risk that our system may inadvertently perpetuate these biases in its generations. This could manifest in various forms, such as gender, cultural, or industry-specific biases, and could affect the fairness and neutrality of the job postings generated. 
Furthermore, if biased postings are used extensively, they could adversely influence downstream tasks. For example, biased job postings could skew job recommendation algorithms, leading to unfair job suggestions that do not treat all individuals equally. This highlights the need for careful consideration and mitigation of biases in our approach to ensure equitable outcomes in all applications. 

\paragraph{Subset of the Taxonomy.}
Due to limited resources, we restricted the generation of our synthetic dataset to $\sim$8K samples, with a fraction of the ESCO taxonomy that is also used in the \textsc{SkillSpan-M} dataset.
Consequently, the multi-class classifier is also trained to classify with a limited set of skills. Scaling up to the full taxonomy might modify the behavior of the supervised classification model, while it should have little to no impact on the ICL skill-matching pipeline.

\section{Ethics Statement}

In this work, we strictly used publicly available data and generated synthetic datasets, avoiding the use of sensitive or private information. This approach aligns with ethical standards concerning data privacy and security.

However, our system can be used to extract information from personal documents, or be used for sensitive applications in the human resources domain, notably pre-selecting candidates to hire. It shall not be used without the supervision of a human. In this work, we focus on the development of a framework to reduce reliance on real-world annotated data. Extended to resumes, it could allow users to perform the skill extraction and matching task without requiring personal data to be anonymized. Given the limited performance of anonymization tools, generating data following similar distribution would greatly reduce privacy issues for such applications.

\section*{Acknowledgements}

We thank Jibril Frej (EPFL) for fruitful discussions and feedback on the first version of the paper. MZ is supported by the Independent Research Fund Denmark (DFF) grant 9131-00019B and in parts by ERC Consolidator Grant DIALECT 101043235. We also gratefully acknowledge the support of the Swiss National Science Foundation (No. 215390), Innosuisse (PFFS-21-29), the EPFL Science Seed Fund, the EPFL Center for Imaging, Sony Group Corporation, and the Allen Institute for AI.

\bibliography{anthology,custom}


\appendix
\section{Prompts}
\label{sec:app-prompt}

\subsection{Extraction Demonstrations}
\label{sec:extr-prompt-shots}
In our in-context learning pipeline, we provide seven demonstrations to guide the LLM in performing extractions. Below is one example.
\begin{quote}
    \vspace{-2em}
    \itemsep.5em
    \item[] \texttt{\textbf{Sentence}: we are looking for a team leader with strong communication skills to foster collaboration and information sharing within the team.\\
    \textbf{Answer}: We are looking for a team leader with strong @@communication skills\#\# to foster collaboration and information sharing within the team.}
    \item[] \texttt{\textbf{Sentence}: the ability to work collaboratively across disciplines is a key criterion for this position. \\
    \textbf{Answer}: @@ability to collaborate across disciplines\#\# is a key criterion for this position.}
    \item[] \texttt{\textbf{Sentence}: As a Java Senior Software Engineer with experience, you will be a member of a Scrum team. \\
    \textbf{Answer}: As a Java Senior Software Engineer with experience, you will be a member of a Scrum team.}
    \item[] \texttt{\textbf{Sentence:} In her role as a team leader, she has continuously supported the professional development of her employees.\\
    \textbf{Answer}: In her role as a team leader, she has continuously fostered the professional @@development of her employees\#\#.}
    \item[] \texttt{\textbf{Sentence:} He is a resilient employee who has been able to set proper priorities and organize tasks thoughtfully during periods of heavy workload. \\
    \textbf{Answer}: He is a resilient employee who has been able to set @@correct priorities and organize tasks thoughtfully\#\# during periods of high workload.}
    \item[] \texttt{\textbf{Sentence:} Highly qualified, flexible employees from the insurance and IT industry develop them further. \\
    \textbf{Answer}: Highly qualified, flexible employees from the insurance and IT industries continue to develop them.}
    \item[] \texttt{\textbf{Sentence:} Over the past few years, it has succeeded in continuously developing itself in a rapidly changing environment. \\
    \textbf{Answer}: Over the past few years, he has succeeded in @@continuously developing\#\# himself in a rapidly changing environment\#\#.}

\end{quote}

\subsection{Matching}
\label{sec:matching-prompt}

\subsubsection{Prompt}
Here we provide the prompt used to match each extracted skill to one of the pre-selected candidates. The one-shot demonstration used in this prompt is provided in section \ref{sec:matching-prompt-shots}.
\label{sec:matching-prompt-prompt}

\begin{quote}
    \vspace{-2em}
    \itemsep.5em
    \newquote{System}{You are an expert human resource manager. You need to analyse skills in a job posting.}
    \newquote{Instruction}{You are given a sentence from a job description, and a skill extracted from this sentence. Choose from the list of options the one that best match the skill in the context. Answer with the associated letter.}

    \newquote{}{\{Demonstration\}}
    
    \newquote{}{Sentence: \{Sentence\} \\
    Skills: \{Extracted\} \\
    A: \{Candidate 1\} \\
    ...\\
    J: \{Candidate 10\}} 
    \newquote{}{Answer: }
    
\end{quote}

\subsubsection{Demonstration}
The demonstration we use in the matching step (Section \ref{sec:matching-prompt-prompt}) of our in-context learning pipeline.
\label{sec:matching-prompt-shots}

\begin{quote}
    \vspace{-2em}
    \itemsep.5em
    \newquote{Sentence}{Understand basic provisions of copyright and privacy.}\\
    \subquote{Skill}{Data protection.} \\
    \subquote{Options}{\\}
    \subquote{}{A: "Respect privacy principles"} \\
    \subquote{}{B: "Understand data protection"} \\
    \subquote{}{C: "Ensure data protection in aviation operations"} \\
    \subquote{}{D: "Data protection"} \\
    \subquote{Answer}{b, d.}
\end{quote}

\subsection{Generation of dataset}
\label{sec:datagen-prompt}

\subsubsection{Positive samples}
We use this prompt to generate samples containing ESCO skills.
\label{sec:datagen-positive-samples}
\begin{quote}
    \vspace{-2em}
    \itemsep.5em
    \newquote{System}{You are the leading AI Writer at a large, multinational HR agency. You are considered as the world's best expert at expressing required skills and knowledge in a variety of clear ways. You are particularly proficient with the ESCO Occupation and Skills framework. As you are widely lauded for your job posting writing ability, you will assist the user in all job-posting, job requirements and occupational skills related tasks.} 
    \newquote{Instruction}{You work in collaboration with ESCO to gather rigid standards for job postings. Given a list of ESCO skills and knowledges, you're asked to produce a single example of exactly one sentence that could be found in a job ad and refer to all skill or knowledge component. Ensure that your sentence is well written and could be found in real job advertisement. Use a variety of styles. You're trying to provide a representative sample of the many, many ways real job postings would evoke skills. All the skills in : \{skillList\} must be integrated. A candidate should have different degrees of expertise in all the given skills. This degree should be specified for each skills in the sentence. You must not include any skills in ESCO that were not given to you. Try to be as implicit as possible when mentionning the skill. Try not to use the exact skill string \{wordsToAvoid\}. Avoid explicitly using the wording of this extra information in your examples. Your sentence must not start with 'We are seeking', 'We are looking' or 'We are searching'. Generate stricly only one example.}
\end{quote}

\subsubsection{Negative samples}
\label{sec:datagen-negative-samples}
We use two different prompts to generate negative samples:
This first prompt generates negative samples that describe the company.
\begin{quote}
    \vspace{-2em}
    \itemsep.5em
    \newquote{System}{You are the leading AI Writer at a large, multinational HR agency. You are considered as the world's best expert at writing introductions of job posting.}
    \newquote{Instruction}{You are the leading AI Writer at a large, multinational HR agency. You are considered as the world's best expert at writing introductions of job posting. You should write \{nExamples\} examples of the first line of the job posting. It should consists in introducing the company, its localization, the number of employees, and any information relevant to a future candidates who wants to learn about the company. The description should be concise, specify the potential growth of the company and a domain of action. You shouldn't mentoin anything about the actual job, no skills required for the candidate and shouldn't mention the candidate at all. You should mention a wide range of company field, size, and localization in each of the examples.}

\end{quote}

\noindent This second prompt generates sentences detailing the salary and perks of a job.

\begin{quote}
    \vspace{-2em}
    \itemsep.5em
    \newquote{System}{You are the leading AI Writer at a large, multinational HR agency. You are considered as the world's best expert at specifying administrative information in job posting.}
    \newquote{Instruction}{You are the leading AI Writer at a large, multinational HR agency. You are considered as the world's best expert at specifying administrative information in job posting. You should produce \{nExamples\} descriptions of the salary and the perks a candidate to a certain job would have. You shouldn't mention the actual job and the candidate itself. You could add diversity by varying the salary and the perks. You must write a salary range between 40k and 100k according to the job in half of your generation.}
\end{quote}

\subsection{Refinement of dataset}
\label{sec:datagen-refined-prompt}

\subsubsection{Initial prompt}
\label{sec:datagen-initial-prompt}

\begin{quote}
    \vspace{-2em}
    \itemsep.5em
    \newquote{System}{You are an expert human resource manager. You need to analyse skills in a job posting.}
    \newquote{Instruction}{You are an expert human resource manager. You are given an extract from a job description and a skill coming from ESCO. Highlight all the parts of the job description that relates to the given skill, by surrounding them with tags '@@' at the beginning and '\#\#' at the end. You should rewrite the entire sentence. The highlighted parts should precisely talk about the given skills and only this skills. The higlighted parts must precisely be about the given skills. Do not highlight parts not related to it. The sentence should be rewritten perfectly, using the same exact same words. You must highlight at least one part in the sentence that you will rewrite. The highlighted part should be as short as possible.}

\end{quote}

\subsubsection{Refining shots}
\label{sec:datagen-refining-shots}

In case of incorrectly bound annotations :

\begin{quote}
    \vspace{-2em}
    \itemsep.5em
    \newquote{}{In your response, you highlighted some parts using @@ at the beginning and @@ at the end. Please use @@ at the beginning of the parts and \#\# at the end of the part you want to highlight. Annotate the previous sentence, but with the correct highlighting.}
\end{quote}

When there is a lack of annotations : 

\begin{quote}
    \vspace{-2em}
    \itemsep.5em
   \newquote{}{In your response, you highlighted nothing. Please annotate the previous sentence, and highlight at least one part linked to the skill.}
\end{quote}

\section{Ablation studies - Few-Shot ICL}
\label{sec:app-ablation}

\subsection{Demonstrations}
To conduct the experiments for the In-context Learning with LLM, we will use the demonstrations retrieval from the training set to provide few shots for both the extraction and the matching. We need to determine the number of demonstration to use for both parts. For this purpose we conduct an ablation study on \textsc{SkillSpan-M} 's validation test trying different configuration of number of shots. We try the following experiments :

\begin{itemize}
        \item \textit{baseline} : Same shots for all the sentences \ref{sec:extr-prompt-shots} \ref{sec:matching-prompt-shots}
        \item $M_1$ : 1 demonstration for the matching part, baseline shot for extraction
        \item $E_5$ : 5 demonstration for the extraction part, baseline shot for matching
        \item $E_7$ : 7 demonstration for the extraction part, baseline shot for matching
        \item $E_{10}$ : 10 demonstration for extraction, baseline shot for matching
        \item $E_7M_1$ : 1 demonstration for the matching part and 7 for the extraction part
        \item $E_7M_3$ : 3 demonstrations for the matching part and 7 for the extraction part
    \end{itemize}

\begin{table}[!ht]
    \centering
    \resizebox{\linewidth}{!}{
    \begin{tabular}{cccc}
    \toprule
              & Recall & Precision & $F_1$ \\ \midrule
                baseline & 0.260 & 0.303 & 0.280 \\
                $E_5$ & 0.279 & 0.296 & 0.287 \\
                $E_7$ & 0.282 & 0.301 & 0.291 \\
                $E_{10}$ & 0.282 & 0.298 & 0.289 \\
                $M_1$ & 0.267 & \textbf{0.305} & 0.284 \\
                $E_7M_1$ & \textbf{0.289} & 0.298 & \textbf{0.2934} \\
                $E_{10}M_3$ & 0.283 & 0.293 & 0.288 \\
        \bottomrule
    \end{tabular}
    }
    \caption{Ablation study for In-context Learning: Selecting optimal number of demonstrations for extraction and matching with GPT-3.5}
    \label{tab:app-ablation_ndem}
\end{table}

\begin{figure}[!ht]
    \centering
    \includegraphics[width=0.5\textwidth]{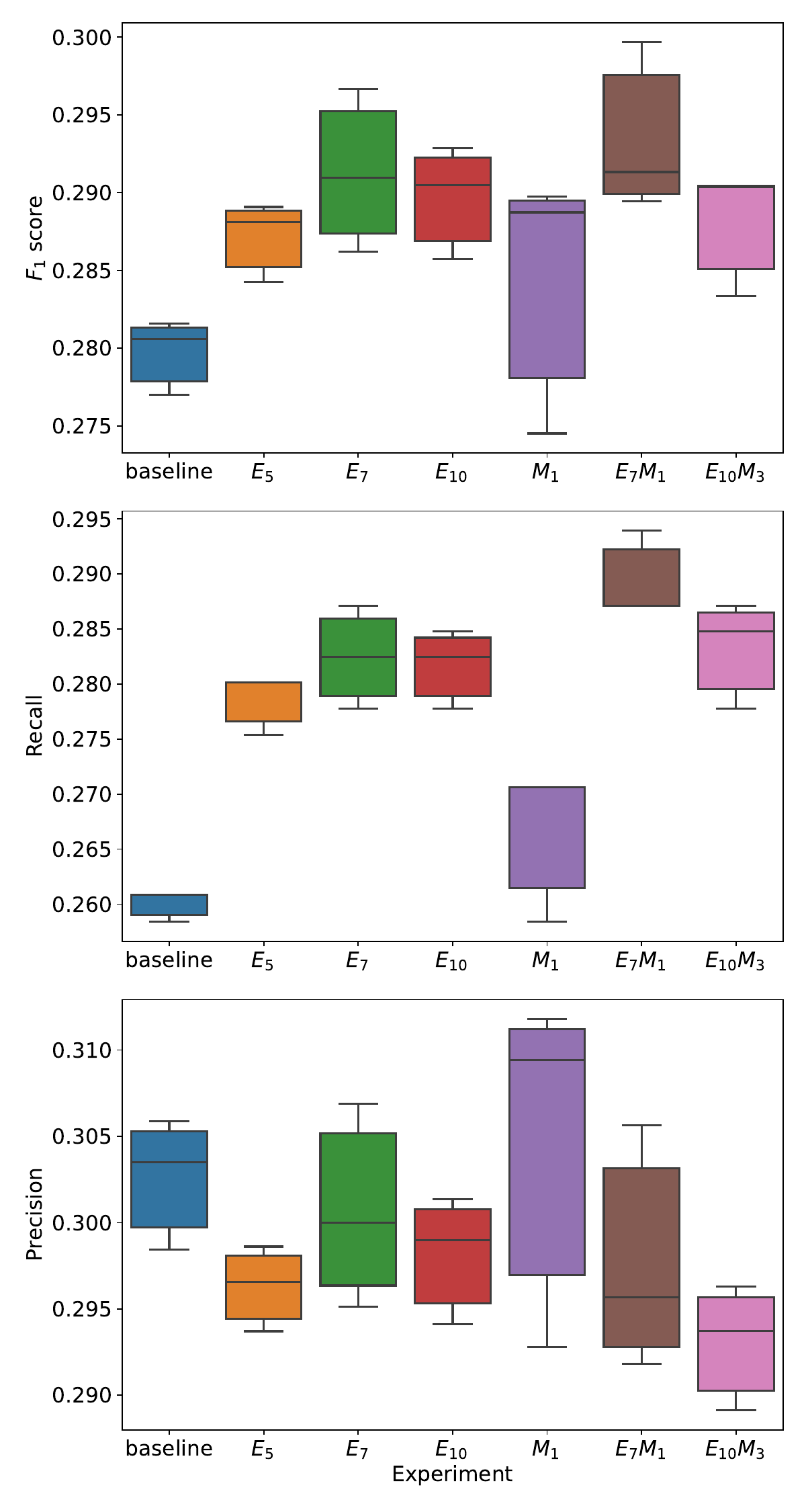}
    \caption{Ablation study for In-context Learning}
    \label{fig:app-ablation_ICL}
\end{figure}

Given the stats in Table \ref{tab:app-ablation_ndem}, displayed on Figure \ref{fig:app-ablation_ICL} we see the adding the demonstration retrieval for the extraction part yields a significative improvement on the recall. We will run the subsequent experiments with 7 demonstrations for the extraction part and one demonstration for the matching part.

\subsection{Candidate Selection}
\begin{figure}[t]
    \centering
    \includegraphics[width=.48\textwidth]{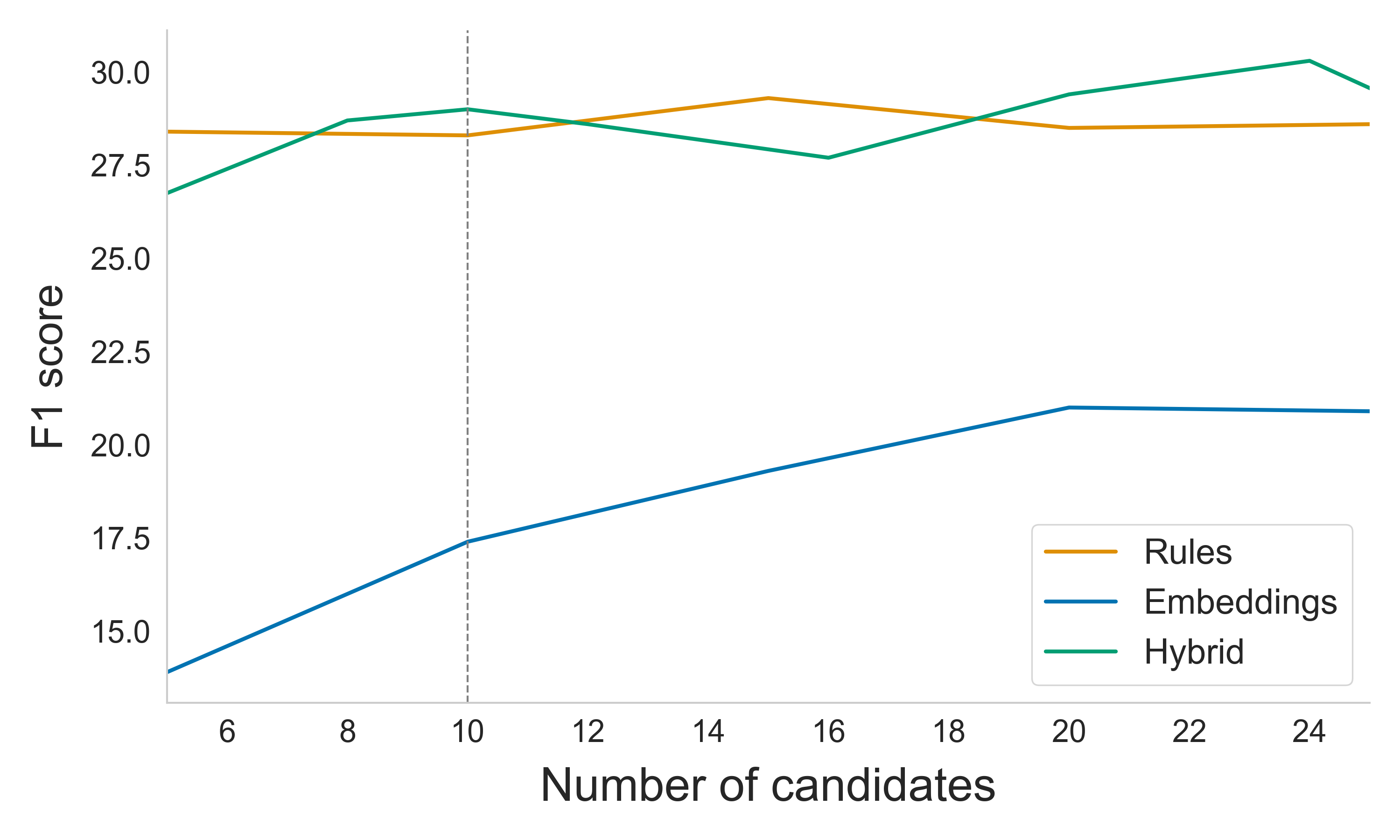}
    \caption{Rule-based, embedding-based, and hybrid candidate selection methods to select $n$ candidates. Note, since the hybrid method takes the union of rule-based and embedding-based methods, $n=5$ using the hybrid method would approximate $n \times 2$ actual number of actual candidates selected}
    \label{fig:num_cand}
\end{figure}

Figure \ref{fig:num_cand} shows shows the $F_1$ scores of the ICL when we vary the number of candidates selected using the rule-based, embedding-based, and hybrid candidate selection methods, holding other elements constant. Looking at our results, we elect to use 10 candidates ($n=5$) with the hybrid method. 

While further increasing the candidates can increase matching performance slightly, we find that providing too many candidates can lead to a noticeable increase in inference time.

\subsection{Matching Step}\label{sec:app-ablation-matching}
We conduct an ablation study on the matching. We remove the matching step from the pipeline and we only extract the spans from the inputted sentences and use rule-based to find matches. We focus on the rule-based method that yields the best results when extracting a small amount of candidates. 
Table \ref{tab:ablation-study-matching} shows that the top-selected candidates are behind the performance of using GPT-3.5 as a re-ranker by around 8\%. Therefore, we continue our experiments using the full three-step pipeline.

\begin{table}[t]
    \centering
    \resizebox{.8\linewidth}{!}{
    \begin{tabular}{crrr}
        \toprule
        \# of Candidates & Precision & Recall & $F_1$ \\
        \midrule
         1 & 24.2 & 16.9 & 19.9 \\
         2 & 17.1 & 23.5 & 19.8 \\
         \bottomrule
    \end{tabular}}
    \caption{Ablation study of the matching step: Performance of the ICL pipeline when taking only the top 1 or 2 candidates using the rule-based selection methods.}
    \label{tab:ablation-study-matching}
\end{table}

\section{Qualitative Analysis}
We present various qualitative examples of predictions from the test set of \textsc{SkillSkape} in Table \ref{tab:quali}. These examples include outcomes from both the supervised multi-label classifier and the in-context learning results using GPT-4. A key observation is the relatively lower number of skills predicted by the supervised classifier, which operates with a threshold of 0.15. Generally, these predictions are feasible and align closely with the gold standard label. However, it should be noted that the evaluation process tends to penalize these predictions for their limited scope.

\newcolumntype{b}{X}
\newcolumntype{s}{>{\hsize=.6\hsize}X}
\begin{table*}[t]
    \centering
    \small
    \resizebox{\linewidth}{!}{
    \begin{tabularx}{\linewidth}{bsss}
    \toprule
    \textbf{Sentence}   &    \textbf{Multi-label Classifier}      &   \textbf{In-context Learning}  & \textbf{Gold}     \\
    \midrule
    (1) Seeking a highly skilled individual with extensive expertise in overseeing and optimizing the operation and maintenance of various technical components and systems on board maritime vessels.                & shipping industry  & overseeing and optimizing the operation and maintenance of various technical components and systems on board maritime vessels & manage vessel engines and systems \\    \midrule
    (2) As an integral part of our team, the ideal candidate should possess a deep understanding of coordinating the alignment and seamless interaction of various system components, while executing rigorous testing and implementing an overarching strategy for the integration of ICT systems & ICT system integration, define integration strategy & coordinating the alignment and seamless interaction of various system components, rigorous testing, integration of ICT systems & ICT system integration, define integration strategy, define software architecture, manage ICT data architecture \\ \midrule
    (3) Ability to effectively adapt to changing circumstances while maintaining a vigilant attitude, maintaining composure in challenging situations, and efficiently managing workload and responsibilities. & handle stressful situations & effectively adapt to changing circumstances, vigilant attitude, composure, efficiently managing workload and responsibilities & exercise patience, adjust priorities, stay alert \\ \midrule
    (4) Are you an experienced professional with a proven track record in designing and implementing comprehensive technology testing frameworks, ensuring the seamless integration of software applications and systems? & develop ICT test suite, execute software tests & designing and implementing comprehensive technology testing frameworks, seamless integration of software applications and systems & develop ICT test suite \\
    \bottomrule
    \end{tabularx}}
    \caption{
    We show several qualitative examples of predictions on the test set of \textsc{SkillSkape} using the supervised multi-label classifier and in-context learning results with GPT-4.}
    \label{tab:quali}
\end{table*}

\section{Other Summary Statistics on \textsc{SkillSkape}}

\subsubsection{Skill Groups}
\label{skillgroups}

We show in Table \ref{tab:skillgrps} the skill groups and counts of skills in each ESCO skill group that is included in the label spaced used for the \textsc{SkillSkape} dataset.

\begin{table}[!ht]
    \centering
\begin{tabular}{lc}
\toprule
Skill Group & skill count  \\
\midrule
\textbf{agriculture, forestry, fisheries and veterinary} & 4 \\
\textbf{arts and humanities} &  8 \\
\textbf{assisting and caring} &  13 \\
\textbf{business, administration and law} &  40 \\
\textbf{communication, collaboration and creativity} & 111 \\
\textbf{constructing} &  3 \\
\textbf{education} &  3 \\
\textbf{engineering, manufacturing and construction} & 22 \\
\textbf{generic programmes and qualifications} & 6 \\
\textbf{handling and moving} &  15 \\
\textbf{health and welfare}  & 7 \\
\textbf{information and communication technologies (icts)} &  71 \\
\textbf{information skills} & 57 \\
\textbf{management skills} & 65 \\
\textbf{natural sciences, mathematics and statistics} & 10 \\
\textbf{services} & 5 \\
\textbf{social sciences, journalism and information} & 1 \\
\textbf{working with computers} & 35 \\
\textbf{working with machinery and specialised equipment} & 14 \\
\textbf{TOTAL} & 514 \\
\bottomrule
\end{tabular}
    \caption{ESCO skill groups present in \textsc{SkillSkape} dataset}
    \label{tab:skillgrps}
\end{table}

\subsubsection{Sentence Length}
\begin{figure}[!ht]
    \centering
    \includegraphics[width=0.48\textwidth]{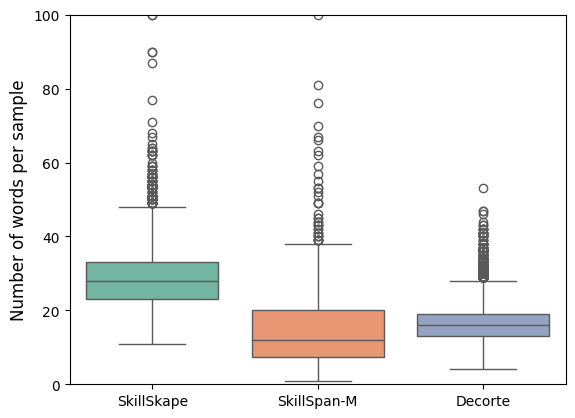}
    \caption{Sentence length distribution in the three datasets. \textsc{SkillSkape} has much longer sentences. \textsc{Decorte} has very short sentences and low length variance.}
    \label{fig:app-sent-len}
\end{figure}

Looking at Figure \ref{fig:app-sent-len}, we can see that \textsc{SkillSkape} has longer sentences and contains more variation in sentence length than \textsc{Decorte}. The distribution of \textsc{SkillSkape} resembles more that of real-world data (\textsc{SkillSpan-M}).

\end{document}